\documentclass[11pt,a4paper]{article}
\usepackage[hyperref]{acl2020}
\pdfoutput=1
\usepackage{graphicx}  
\usepackage{times}
\usepackage{latexsym}

\usepackage[normalem]{ulem}
\usepackage{multirow}
\usepackage{multicol}

\usepackage{url}
\usepackage{microtype}
\usepackage{enumitem} 

\aclfinalcopy 

\title{Efficient strategies for hierarchical text classification: \\External knowledge and auxiliary tasks}

\author{
  Kervy Rivas Rojas, Gina Bustamante, Arturo Oncevay\textsuperscript{\textdaggerdbl}, Marco A. Sobrevilla Cabezudo\textsuperscript{\textdagger}\\
  Research Group on Artificial Intelligence, Pontificia Universidad Cat\'olica del Per\'u, Peru \\
  \textsuperscript{\textdaggerdbl}School of Informatics, University of Edinburgh, Scotland \\
  \textsuperscript{\textdagger}Instituto de Ci\^encias Matem\'aticas e de Computa\c{c}\~ao, Universidade de S\~ao Paulo, Brazil\\
  \texttt{k.rivas@pucp.pe, gina.bustamante@pucp.edu.pe,}\\ 
  \texttt{a.oncevay@ed.ac.uk, msobrevillac@usp.br}\\
}

\date{}

\begin{document}
\maketitle

\begin{abstract}
In hierarchical text classification, we perform a sequence of inference steps to predict the category of a document from top to bottom of a given class taxonomy. Most of the studies have focused on developing novels neural network architectures to deal with the hierarchical structure, but we prefer to look for efficient ways to strengthen a baseline model. We first define the task as a sequence-to-sequence problem. Afterwards, we propose an auxiliary synthetic task of bottom-up-classification. 
Then, from external dictionaries, we retrieve textual definitions for the classes of all the hierarchy's layers, and map them into the word vector space. We use the class-definition embeddings as an additional input to condition the prediction of the next layer and in an adapted beam search. Whereas the modified search did not provide large gains, the combination of the auxiliary task and the additional input of class-definitions significantly enhance the classification accuracy. With our efficient approaches, we outperform previous studies, using a drastically reduced number of parameters, in two well-known English datasets.
\end{abstract}

\section{Introduction}
\label{sec:introduction}

Hierarchical text classification (HTC) aims to categorise a textual description within a set of labels that are organized in a structured class hierarchy~\cite{Silla2011}. The task is perceived as a more challenging problem than flat text classification, since we need to consider the relationships of the nodes from different levels in the class taxonomy~\cite{liu-etal-2019-neuralclassifier}.



Both flat text classification and HTC have been tackled using traditional machine learning classifiers \cite{Liu:2005:SVM:1089815.1089821, Kim:2006:ETN:1175892.1176180} or deep neural networks \cite{Peng:2018:LHT:3178876.3186005, conneau-etal-2017-deep}. Nevertheless, the majority of the latest approaches consider models with a large number of parameters that require extended training time. In the flat-classification scenario, some studies have addressed the problem of efficiency by proposing methods that do not focus on the model architecture, but in external ways of improving the results \cite{joulin-etal-2017-bag, Howard2018}. However, the listed strategies are still underdeveloped for HTC, and the most recent and effective methods are still computationally expensive \cite{yang-etal-2019-deep, banerjee-etal-2019-hierarchical}.


The described context opens our research question: How can we improve HTC at a lower computational cost? Therefore, 
our focus and main contributions are:
\begin{itemize}
    \item A robust model for HTC, with few parameters and short training time, that follows the paradigm of sequence-to-sequence learning.
    \item The practical application of an auxiliary (and not expensive) task that strengthens the model capacity for prediction in a bottom-up scheme.
    \item An exploration of strategies that take advantage of external information about textual definition of the classes. We encode the definitions in the word vector space and use them in: (1) each prediction step and (2) an adapted beam search.
\end{itemize}

\section{Efficient strategies for hierarchical text classification}
\label{sec:htc-improvements}

\subsection{Sequence-to-sequence approach}
\label{sub:nnarchitecture}
Hierarchical classification resembles a multi-label classification where there are hierarchical relationships between labels, i. e.,  labels at lower levels are conditioned by labels at higher levels in the hierarchy. For that reason, we differ from previous work and address the task as a sequence-to-sequence problem, where the encoder receives a textual description and the decoder generates a class at each step (from the highest to the lowest layer in the hierarchy). Our baseline model thereafter is a sequence-to-sequence neural network \cite{NIPS2014_5346} composed of:

\paragraph{Embedding layer:} To transform a word into a vector \textit{w$_{i}$}, where \textit{i} $\in$ \textit{\{1,...,N\}} and \textit{N} is the number of tokens in the input document. We use pre-trained word embeddings from Common Crawl \cite{grave-etal-2018-learning} for the weights of this layer, and we do not fine-tune them during training time.
\paragraph{Encoder:} It is a bidirectional GRU \cite{cho-etal-2014-properties} unit that takes as input a sequence of word vectors and computes a hidden vector \textit{h$_i$} per each \textit{i} time step of the sequence.
\paragraph{Attention layer:}  We employ the attention variant of \citet{DBLP:journals/corr/BahdanauCB14}, and generate a context vector \textit{a$_i$} for each encoder output \textit{h$_i$}.
\paragraph{Decoder:} To use the context \textit{a$_i$} and hidden \textit{h$_i$} vectors to predict the \textit{c$_{l_{jk}}^{l_{j}}$} class of the hierarchy, where \textit{j} $\in$ \textit{\{1,...,M\}}. \textit{M} is the number of levels in the class taxonomy, \textit{l$_j$} represents the j-th layer of the hierarchy, and \textit{l$_{jk}$} is the $k$-th class in level \textit{l$_j$}. Similar to the encoder, we use a bidirectional GRU.

\subsection{Auxiliary task}
\label{sub:multitask-learning}

For an input sequence of words, the model predicts a sequence of classes. Given the nature of recurrent neural networks, iterating over a sequence stores historical information. Therefore, for the last output computation we could take the previous inputs into consideration. 

Previous work in HTC \cite{kowsari2017hdltex, sinha-etal-2018-hierarchical} usually starts by predicting the most general category (Parent node) and continues to a more specific class (Child nodes) each time. However, by following the common approach, the prediction of the most specific classes will have a smaller impact than the more general ones when the error propagates. In this way, it could be harder to learn the relationship of the last target class with the upper ones. 

Inspired by reversing the order of words in the input sequence \cite{NIPS2014_5346}, we propose an auxiliary synthetic task that changes the order of the target class levels in the output sequence. In other words, 
 we go upward from the child nodes to the parent. With the proposed task, the parent and child nodes will have a similar impact on the error propagation, and the network could learn more robust representations.  


\subsection{Class-definition embeddings for external knowledge integration}

We analyze the potential of using textual definitions of classes for external knowledge integration. For each class \textit{c$_{l_{jk}}^{l_{j}}$} in any level $l_{j}$ of the hierarchy, we could obtain a raw text definition from an external dictionary 
to compute a vector representation \textit{cv}, that from now on we call the class definition vector (CDV). We thereafter use the CDV representations with the two following strategies. 

\subsubsection{Parent node conditioning (PNC)}
\label{sub:pc-conditioning}

For a given document \textit{D}, we classify it among the target classes \textit{C = (c$_{l_{1k}}^{l_{1}}$,...,c$_{l_{Mk}}^{l_{M}}$)}, where \textit{M} is the number of layers in the taxonomy. In our approach, we predict the highest-level class \textit{c$_{l_{1k}}^{l_{1}}$} and then use its CDV representation \textit{cv$_{l_{1k}}^{l_{1}}$} as an additional input (alongside the encoder outputs) to the attention layer for the prediction of the next level class \textit{c$_{l_{2k}}^{l_{2}}$}. We continue the process for all the layers of the class hierarchy.  

\subsubsection{Adapted beam search}
\label{sub:beam-search}
Beam search is a search strategy commonly used in neural machine translation \cite{freitag-al-onaizan-2017-beam}, but the algorithm can be used in any problem that involves word-by-word decoding. We assess the impact of applying beam search in HTC, and introduce an adapted version that takes advantage of the computed CDV representations: 

\begin{equation}\label{eq:modified-beam-search}
\sum_{i=0}^{T}logP(y^{i}| x, y^{1},...,y^{t-1}) + CD(z, y^{i})
\end{equation}

In each step of the decoding phase, we predict a class that belongs to the corresponding level of the class hierarchy.  Given a time step \textit{i}, the beam search expands all the $k$ (beam size) possible class candidates and sort them by their logarithmic probability. In addition to the original calculation, we compute the cosine distance between the CDV of a class candidate and the average vector of the word embeddings from the textual description \textit{z} that we want to classify (CD component in Equation \ref{eq:modified-beam-search}). We add the new term to the logarithmic probability of each class candidate, re-order them based on the new score, and preserve the top-$k$ candidates. 

Our intuition behind the added component is similar to the shallow fusion in the decoder of a neural machine translation system \cite{GULCEHRE2017137}. Thus, the class-definition representation might introduce a bias in the decoding, and help to identify classes with similar scores in the classification model. 

\section{Experimental setup}
\label{sec:ex-setup}

\paragraph{Datasets.} We test our model and proposed strategies in two well-known hierarchical text classification datasets previously used in the evaluation of state-of-the-art methods for English: Web of Science \cite[WOS;][]{kowsari2017hdltex} and DBpedia \cite{sinha-etal-2018-hierarchical}. The former includes parent classes of scientific areas such as Biochemistry or Psychology, whereas the latter considers more general topics like Sports Season, Event or Work. General information for both datasets is presented in Table \ref{tab:datasets-info}. 

\begin{table}[t!]
\centering
\begin{tabular}{lrr}
 & WOS & DBpedia \\ \hline
Number of documents & 46,985 & 342,782 \\
Classes in level 1 & 7 & 9 \\
Classes in level 2 & 143 & 70 \\
Classes in level 3 & NA & 219 \\
\hline
\end{tabular}
\caption{\label{tab:datasets-info} Information of WOS and DBPedia corpora}
\end{table}

\paragraph{Model, hyper-parameters and training.} We use the AllenNLP framework \cite{gardner-etal-2018-allennlp} to implement our methods. Our baseline consists of the model specified in \S\ref{sub:nnarchitecture}. For all experiments, we use 300 units in the hidden layer, 300 for embedding size, and a batch size of 100. During training time, we employ Adam optimiser \cite{kingma2014adam} with default parameters ($\beta_{1}=0.9, \beta_{2}=0.98, \varepsilon=10^{-9}$). We also use a learning rate of 0.001, that is divided by ten after four consecutive epochs without improvements in the validation split. Furthermore, we apply a dropout of 0.3 in the bidirectional GRU encoder-decoder, clip the gradient with 0.5, and train the model for 30 epochs. For evaluation, we select the best model in the validation set of the 30 epochs concerning the accuracy metric.

\paragraph{Settings for the proposed strategies.} 
\begin{itemize}
    \item For learning with the auxiliary task, we interleave the loss function between the main prediction task and the auxiliary task (\S\ref{sub:multitask-learning}) every two epochs with the same learning rate. 
    We aim for both tasks 
    to have equivalent relevance in the network training.
    \item To compute the class-definition vectors, we extract the textual definitions using the Oxford Dictionaries API\footnote{\url{https://developer.oxforddictionaries.com/}}. We vectorize each token of the descriptions using pre-trained Common Crawl embeddings (the same as in the embedding layer) and average them. 
    \item For the beam search experiments, we employ a beam size \textit{(k)} of five, and assess both the original and adapted strategies. We note that the sequence-to-sequence baseline model use a beam size of one\footnote{In preliminary experiments, we considered a beam size of ten, but we did not note a significant improvement.}.
\end{itemize}

\section{Results and discussion}
\label{sec:resuts}

\begin{table*}[ht!]
\begin{center}
\setlength\tabcolsep{3.5pt}
\resizebox{\linewidth}{!}{%
\begin{tabular}{l|lcc}
\hline
 & \textbf{} & \multicolumn{1}{c}{WOS} & \multicolumn{1}{c}{DBpedia} \\ 
 \hline \hline
\multirow{5}{*}{Individual strategies}
     & seq2seq baseline         & 78.84 $\pm$ 0.17  & 95.12  $\pm$ 0.01\\
     & Auxiliary task           & $^*$78.93 $\pm$ 0.52  & $^*$95.21  $\pm$ 0.16\\
     & Parent node conditioning (PNC) & $^*$79.01 $\pm$ 0.18 & $^*$95.26  $\pm$ 0.09\\
     & Beam search (original)   & $^*$78.90 $\pm$ 0.25 & $^*$95.25  $\pm$ 0.01\\
     & Beam search (modified)   & $^*$78.90 $\pm$ 0.28 & $^*$95.26  $\pm$ 0.01\\ \hline
\multirow{5}{*}{Combined strategies}
     & Auxiliary task + PNC [7M params.] & $^*$79.79 $\pm$ 0.45 & $^*$95.23  $\pm$ 0.13\\
     & Beam search (original) + PNC & $^*$79.18 $\pm$ 0.19 & $^*$\textbf{95.30}  $\pm$ 0.10\\
     & Beam search (modified) + PNC & $^*$79.18 $\pm$ 0.23 & $^*$95.30  $\pm$ 0.11\\
     & Auxiliary task + PNC + Beam search (orig.) & $^*$\textbf{79.92} $\pm$ 0.51 & $^*$95.26  $\pm$ 0.12\\
     & Auxiliary task + PNC + Beam search (mod.) & $^*$79.87 $\pm$ 0.49 & $^*$95.26  $\pm$ 0.12\\ \hline
\multirow{2}{*}{Previous work} 
    & HDLTex~\cite{kowsari2017hdltex} [5B params.] & 76.58 &  92.10 \\
    & \citet{sinha-etal-2018-hierarchical} [34M params.] & 77.46 & 93.72 \\ \hline
\end{tabular}
}
\end{center}
\caption{Test accuracy ($\uparrow$ higher is better) for our proposed strategies, tested separately and combined, and a comparison with previous classifiers. Reported values are averaged across five runs, and $^*$ indicates Almost Stochastic Dominance~\cite{dror-etal-2019-deep} over the seq2seq baseline with a significance level of 0.05. The amount of parameters of each combined strategies is up to seven million.}
\label{tab:ind-results}
\end{table*} 

Table~\ref{tab:ind-results} presents the average accuracy results of our experiments with each proposed method over the test set. For all cases, we maintain the same architecture and hyper-parameters in order to estimate the impact of the auxiliary task, parent node conditioning, and the beam search variants independently. Moreover, we examine the performance of the combination of our approaches\footnote{We tried all the possible combinations, but only report the ones that offer an improvement over the individual counterparts.}. 

In the individual analysis, we observe that the parent node conditioning and the auxiliary task provides significant gains over the seq2seq baseline, which support our initial hypothesis about the relevance of the auxiliary loss and the information of the parent class. 
Conversely, we note that the modified beam search strategy has the lowest gain of all the experiments in WOS, although it provides one of the best scores for DBpedia. One potential reason is the new added term for the $k$-top candidates selection (see Eq. \ref{eq:modified-beam-search}), as it strongly depends on the quality of the sentence representation. The classes of WOS includes scientific areas that are usually more complex to define than the categories of the DBpedia database\footnote{Averaging words vectors to generate a sentence embedding is an elemental approach. 
Further work could explore the encoding of the class-definition embeddings directly from the training data, or to weight the scores of the classification model and the similarity score to balance the contribution of each term.}. 

We also notice that the accuracy increment is relatively higher for all experiments on the WOS corpus than on DBpedia. A primary reason might be the number of documents in each dataset, as DBpedia contains almost seven times the number of documents of WOS. If we have a large number of training samples, the architecture is capable of learning how to discriminate correctly between classes only with the original training data. However, in less-resourced scenarios, our proposed approaches with external knowledge integration could achieve a high positive impact.

As our strategies are orthogonal and focus on different parts of the model architecture, we proceed to combine them and assess their joint performance. In the case of WOS, we observe that every combination of strategies improves the single counterparts, and the best accuracy is achieved by the merge of the auxiliary task and PNC, but with an original beam search of size five. Concerning DBpedia, most of the results are very close to each other, given the high accuracy provided since the seq2seq baseline. However, we note the relevance of combining the PNC strategy with the original or modified beam search to increase the performance.


Finally, we compare our strategies to the best HTC models reported in previous studies \cite{kowsari2017hdltex, sinha-etal-2018-hierarchical}. We then observe that the results of our methods are outstanding in terms of accuracy and number of parameters. Moreover, the training time of each model takes around one hour (for the 30 epochs), and the proposed auxiliary task do not add any significant delay.

\section{Related work}
\label{sec:related-work}

Most of the studies for flat text classification primarily focus on proposing a variety of novel neural architectures \cite{conneau-etal-2017-deep, NIPS2015_5782}. Other approaches involve a transfer learning step to take advantage of unlabelled data. \citet{mccann2017learned} used the encoder unit of a neural machine translation model to provide context for other natural language processing models, while \citet{Howard2018} pre-trained a language model on a general-domain monolingual corpus and then fine-tuned it for text classification tasks.

In HTC, there are local or global strategies \cite{Silla2011}. The former exploits local information per layer of the taxonomy, whereas the latter addresses the task with a single model for all the classes and levels. Neural models show excellent performance for both approaches \cite{kowsari2017hdltex, sinha-etal-2018-hierarchical}. Furthermore, other studies focus on using transfer learning for introducing dependencies between parent and child categories \cite{banerjee-etal-2019-hierarchical} and deep reinforcement learning to consider hierarchy information during inference \cite{mao-etal-2019-hierarchical}.

The incorporation of external information in neural models has offered potential in different tasks, such as in flat text classification. By using categorical metadata of the target classes \cite{kim-etal-2019-categorical} and linguistic features at word-level \cite{margatina-etal-2019-attention}, previous studies have notably improved flat-text classification at a moderate computational cost. Besides, \citet{Liu:2016:RNN:3060832.3061023} outperform several state-of-the-art classification baselines by employing multitask learning.

To our knowledge, the latter strategies are not explicitly exploited for HTC. For this reason, our study focuses on the exploration and evaluation of methods that enable hierarchical classifiers to achieve an overall accuracy improvement with the least increasing complexity as possible.

\section{Conclusion}
\label{sec:conclussions}
We presented a bag of tricks to efficiently improve hierarchical text classification by adding an auxiliary task of reverse hierarchy prediction and integrating external knowledge (vectorized textual definitions of classes in a parent node conditioning scheme and in the beam search). Our proposed methods established new state-of-the-art results with class hierarchies on the WOS and DBpedia datasets in English. Finally, we also open a path to study integration of knowledge into the decoding phase, which can benefit other tasks such as neural machine translation.

\section*{Acknowledgements}
We are thankful to the Informatics' support team at PUCP, and specially to Corrado Daly. We also appreciate the collaboration of Robert Aduviri and Fabricio Monsalve in a previous related project that build up our research question. Besides, we thanks the comments of Fernando Alva-Manchego on a draft version and the feedback of our anonymous reviewers. Finally, we acknowledge the support of NVIDIA Corporation with the donation of a Titan Xp GPU used for the study.

\bibliographystyle{acl_natbib}
\bibliography{acl2020}

\end{document}